\documentclass[runningheads]{llncs}

\usepackage[mobile]{eccv}

\usepackage{eccvabbrv}

\usepackage{graphicx}
\usepackage{booktabs}

\usepackage[accsupp]{axessibility} 

\usepackage[pagebackref,breaklinks,colorlinks,citecolor=blue,linkcolor=blue]{hyperref}

\usepackage{multirow}
\usepackage{rotating}
\usepackage{array}
\usepackage{tabularx}
\usepackage{graphicx}
\usepackage{subcaption}
\usepackage{float}
\usepackage{wrapfig}
\usepackage{mdframed}

\newcommand{\modelname}{\textit{CE3D}}

\begin{document}

\title{Chat-Edit-3D: Interactive 3D Scene Editing \\ via Text Prompts} 

\author{Shuangkang Fang\inst{1} \and
Yufeng Wang\inst{1} \and
Yi-Hsuan Tsai\inst{2} \and
Yi Yang \inst{3} \and
Wenrui Ding \inst{1} \and
Shuchang Zhou \inst{3} \and
Ming-Hsuan Yang \inst{2,4}
}

\authorrunning{SK. Fang et al.}

\institute{$^1$ Beihang University~  $^2$ Google~ $^3$ Megvii~ $^4$ University of California, Merced \\
\url{https://sk-fun.fun/CE3D}
}

\maketitle

\begin{abstract}
Recent work on image content manipulation based on vision-language pre-training models has been effectively extended to text-driven 3D scene editing.
However, existing schemes for 3D scene editing still exhibit certain shortcomings, hindering their further interactive design.
Such schemes typically adhere to fixed input patterns, limiting users' flexibility in text input. Moreover, their editing capabilities are constrained by a single or a few 2D visual models and require intricate pipeline design to integrate these models into 3D reconstruction processes.
To address the aforementioned issues, we propose a dialogue-based 3D scene editing approach, termed \modelname{}, which is centered around a large language model that allows for arbitrary textual input from users and interprets their intentions, subsequently facilitating the autonomous invocation of the corresponding visual expert models.
Furthermore, we design a scheme utilizing Hash-Atlas to represent 3D scene views, which transfers the editing of 3D scenes onto 2D atlas images.
This design achieves complete decoupling between the 2D editing and 3D reconstruction processes, enabling \modelname{} to flexibly integrate a wide range of existing 2D or 3D visual models without necessitating intricate fusion designs.
Experimental results demonstrate that \modelname{} effectively integrates multiple visual models to achieve diverse editing visual effects, possessing strong scene comprehension and multi-round dialog capabilities.


\end{abstract}

\section{Introduction} \label{sec:intro} 
\begin{figure}[t]
    \centering
    \includegraphics[width=1.\textwidth]{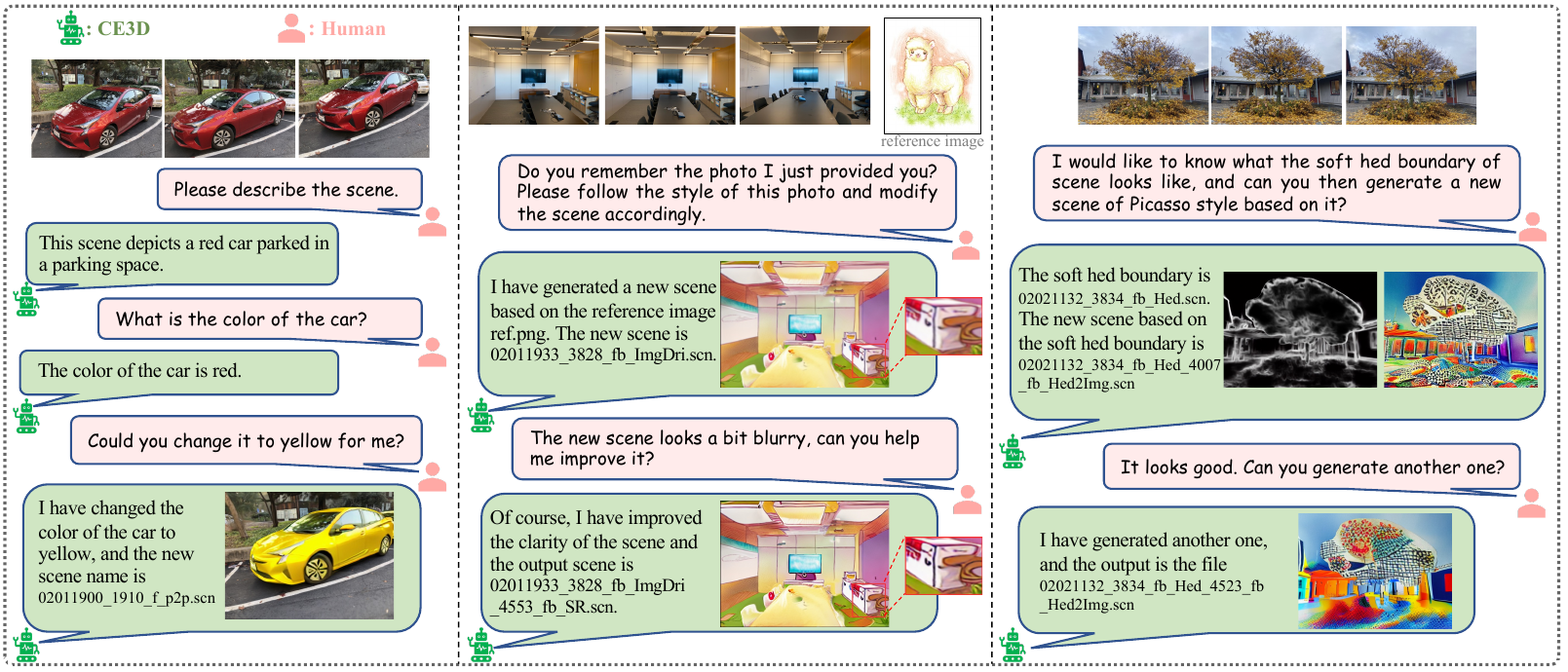}
    \caption{\textbf{Examples of chatting with \modelname{}}. We propose \modelname{}, a novel paradigm for 3D scene editing,
    which is compatible with a variety of extant visual models. %
    By managing these visual experts through the LLMs, we achieve challenging dialogue-based scene editing tasks that are difficult to accomplish with previous methods.}
    \label{fig:shocking}
    \vspace{-8pt}
\end{figure}

The emergence of advanced conversational systems such as ChatGPT~\cite{chatGPT3.5,achiam2023gpt-4}, based on Large Language Models (LLMs)~\cite{brown2020gpt3, Driess2023_PaLM,touvron2023_llama,wei2022chain-1}, signifies a new stage in human-computer interaction. These systems, trained on massive datasets, possess the ability to comprehend complex contexts, engage in logical reasoning, and generate coherent responses. They have been used to address practical issues while significantly reducing human effort.
In addition to text processing capabilities, recent research has successfully applied LLMs to the 2D visual domain. Examples such as Visual-ChatGPT~\cite{wu2023visual-gpt} and MM-REACT~\cite{yang2023mm-react} leverage LLMs as a logical hub, facilitating the integration and management of multiple visual models like detection, segmentation, visual question answering (VQA), etc., while achieving a more natural interaction with users.

\begin{figure}[ht]
    \centering
    \includegraphics[width=0.7\textwidth]{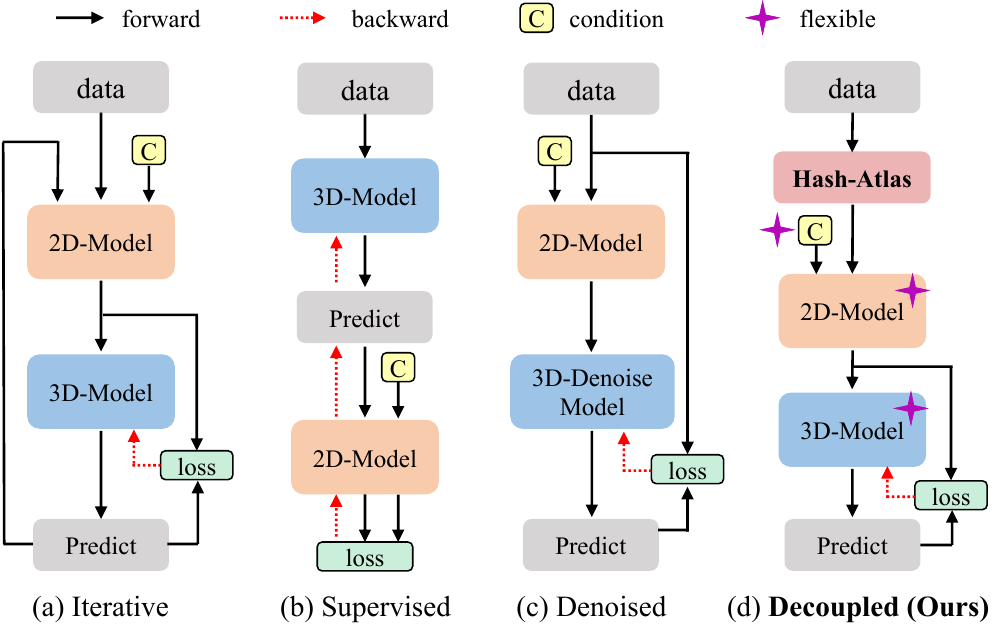}
    \caption{\textbf{Differences between \modelname{} and other typical text-driven editing frameworks}. Existing methods (a)-(c) require complex design to integrate specific 2D and 3D models. While \modelname{} (d) decouples the 2D editing process from the 3D representation, enhancing its compatibility with various visual models.}
    \label{fig:diff_pipe}
    \vspace{-5pt}
\end{figure}

Motivated by the success of LLMs, we exploit LLMs for 3D scene editing in this work (see Fig.~\ref{fig:shocking}).
Nevertheless, constructing such a dialogue system is challenging primarily because open-ended dialogue implies diverse user queries and demands, in which a single or a few editing approaches cannot adequately address. One potential solution is to integrate multiple visual editing methods; however, existing text-driven 3D editing frameworks, such as those based on NeRF~\cite{mildenhall2020nerf,haque2023instruct,fang2023DN2N,wang2022clipnerf,wang2022nerfart,yin2023or-nerf} or Gaussian-splatting~\cite{kerbl20233dgaussian,fang2023gaussianeditor,ye2023gaussian-seg-edit}, constrain this development. We summarize these typical frameworks in Fig.~\ref{fig:diff_pipe}:
(a) iterative methods like IN2N~\cite{haque2023instruct} and Gaussian-Editor~\cite{fang2023gaussianeditor} require iteratively embedding of information from 2D visual models into 3D scenes. (b) Supervised approaches~\cite{yin2023or-nerf,wang2022clipnerf} rely on 2D models to supervise the reconstruction process. (c) Denoised method DN2N~\cite{fang2023DN2N} necessitates the prior modeling of inconsistency in the 2D editing model. 
These coupled designs are often intricate and require identifying well-suited 2D and 3D models that function effectively within such frameworks. This makes it difficult to flexibly integrate diverse visual models, thus hindering the deployment of LLMs as well as resulting in limited editing capabilities and constraints imposed by the textual forms of 2D models.

Based on the above discussion, we propose a novel editing paradigm by fully decoupling the 2D model from the 3D reconstruction process, as shown in Fig.~\ref{fig:diff_pipe}\textcolor{blue}{(d)}. To achieve this, a neural network that maps the different views of a 3D scene to plane atlases is initially learned. Subsequently, it transforms the editing of the 3D scene into operations performed within the 2D atlas space. Since the atlases inherently accommodate 2D visual models, there is no need for additional design to adapt 2D visual models to the 3D reconstruction process, which ensures the flexible integration of multiple visual experts. Finally, by leveraging LLMs for parsing user text input and managing various visual models, our framework achieves editing of a 3D scene through chatting, which we refer to as Chat-Edit-3D (\modelname{}).

The main contributions of our paper are as follows:
\begin{enumerate}
    \item We develop a mapping network called Hash-Atlas, which 
    translates the editing of 3D scenes into the manipulation of 2D atlases, thus decoupling the editing and reconstruction processes to avoid the intricacies of conventional pipeline architectures.

    \item By employing the LLMs, we propose a dialogue framework for editing 3D scenes, termed \modelname{}, which encompasses a mechanism for parsing user queries and formulating user responses, as well as managing multiple visual models and scene files. Moreover, it incorporates a purpose-built executor to coordinate the processing of atlases, thereby enabling the editing of scenes.
    
    \item Experimental results demonstrate that \modelname{} exhibits strong scalability and is compatible with various existing 2D and 3D visual models. Compared to previous methods, \modelname{} possesses more robust text parsing capabilities, richer editing capabilities, and more natural interaction.
\end{enumerate}

\section{Related Work} \label{sec:related_works}
\noindent{\textbf{Vision-Language Pretrained Models.}}
The advances in large-scale vision-language pre-training methods~\cite{radford2021clip,li2022blip,li2023blip2,dhariwal2021diffusion,rombach2022high,nichol2021glide,kawar2022imagic,ho2020denoising,saharia2022photorealistic,ho2022classifier} have endowed models with the capability to comprehend rich visual and textual information. 
These pre-trained models are currently widely employed to boost the performance of downstream tasks. For example, combining CLIP with diffusion~\cite{dhariwal2021diffusion,ho2020denoising,song2020denoising,saharia2022photorealistic} allows for high-fidelity image generation based on text~\cite{rombach2022high,ho2022classifier}.
More relevant to our study are controllable image generation and editing methods. For instance, ControlNet~\cite{zhang2023controlnet} can generate new images based on conditions like text, depth, and edge. Instruct-P2P~\cite{brooks2022instructpix2pix} facilitates instruction-based image editing, and BLIP-diffusion~\cite{li2023blip-diffusion} enables style transfer based on reference images.

\vspace{1mm}
\noindent{\textbf{LLMs for Vision Tasks.}}
The emergence of LLMs~\cite{brown2020gpt3,chatGPT3.5,achiam2023gpt-4, Driess2023_PaLM,touvron2023_llama} showcases the potent capabilities of a new generation of language models, providing a robust foundation for the research and application of dialogue systems. 
However, LLMs lack the inherent ability to process visual inputs directly. One solution involves mapping visual information into text~\cite{hu2022-l2v-1,wang2022-l2v-2,yang2022empirical-l2v-3,zeng2022socratic-l2v-4}, allowing subsequent processing by LLMs.
Another solution~\cite{wu2023visual-gpt,yang2023mm-react,suris2023vipergpt} leverages LLMs' chain-of-thought capability~\cite{wei2022chain-1,brown2020gpt3}, treating LLMs as an inference hub and calling upon relevant visual experts based on the user's text input to handle various visual tasks. 
There are also some methods of applying LLMs to 3D scenes, such as Chat-3D~\cite{wang2023chat3D} and 3D-GPT~\cite{sun20233d-gpt}. Chat-3D primarily focuses on interpreting scene content and addressing tasks such as VQA and image captioning, while 3D-GPT emphasizes utilizing LLMs for the generation of 3D scenes. In this paper, we are mainly concerned with exploring how LLMs can assist in editing 3D scenes.

\vspace{1mm}
\noindent{\textbf{3D Scene Editing.}}
3D scene editing involves both the processes of scene reconstruction and editing. Earlier methods were based on structural representation like point clouds and meshes~\cite{huang2021learning,mu20223d,zhou2014color,hollein2022stylemesh,han2021exemplar}. Recently, scene editing techniques based on NeRF~\cite{mildenhall2020nerf} have emerged.
Examples include leveraging the implicit modeling nature of NeRF for editing~\cite{martin2021nerfinthewild, kobayashi2022decomposing,tang2022CCNeRF,li2022climatenerf}, employing structural transformations to achieve diverse editing capabilities~\cite{fang2022pvd,fang2023pvdal}, and utilizing images as references for style transfer~\cite{huang2022stylizednerf,gu2021stylenerf,zhang2022arf,boss2021nerd,boss2021neural,mu20223dphotostylization}.
The integration with pre-trained text-visual models has yielded impressive results in 3D scene editing research, as seen in works like CLIP-NeRF~\cite{wang2022clipnerf}, IN2N~\cite{haque2023instruct}, Nerf-Art~\cite{wang2022nerfart}, DN2N~\cite{fang2023DN2N}, among others. Beyond NeRF, the Gaussian-splatting method~\cite{kerbl20233dgaussian} has also demonstrated potential in 3D scene editing~\cite{fang2023gaussianeditor,ye2023gaussian-seg-edit}.
However, these methods heavily depend on intricate pipelines combining 3D reconstruction with 2D pretrained models, complicating component modification and replacement. Additionally, their rigidity in input modes hinders adaptation to diverse text expressions.

\begin{figure*}[t]
    \centering
    \includegraphics[width=1.\textwidth]{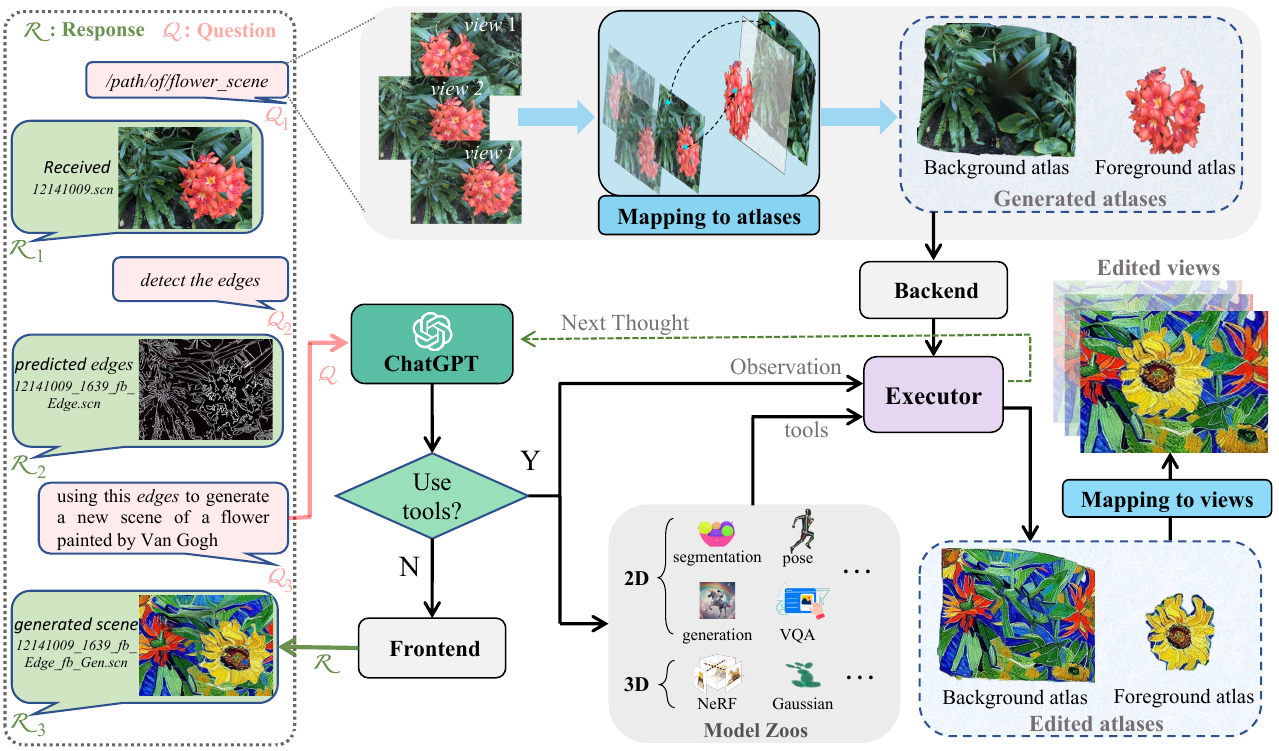}
    \caption{\textbf{Illustration of the \modelname{} framework}. The basic process is as follows: (1) Given the user's text query, ChatGPT interprets the text and determines whether visual tools are required for this dialogue.
    (2) When visual tools are needed, ChatGPT will call the desired tools from the model zoo and provide them with corresponding parameters. 
    (3) The Backend further queries the atlases and other files to be invoked. In addition, if the atlases do not exist, the Backend first acquires them using the Hash-Atlas network.
    (4) The Executor executes visual tools to edit the atlases and feeds back new status to ChatGPT for subsequent actions. The edited atlases are then mapped back to the 3D scene views through the Hash-Atlas network for later scene reconstruction.
    (5) Since one dialogue may require multiple model calls, ChatGPT repeats the above process until it determines that visual tools are no longer needed. Then the Frontend responds to the user with the editing results and ChatGPT's outputs.
    }
    \label{fig:overview_pipe}
\end{figure*}

\section{Proposed Method} \label{sec:method}
In this section, we first illustrate the \modelname{} pipeline (Fig.~\ref{fig:overview_pipe}), followed by the design of the Hash-Atlas network (Sec~\ref{subsec:hash-atlas}), editing strategy in atlas space (Sec~\ref{subsec:editing_atlas}), and components of dialogue system within \modelname{} (Sec~\ref{subsec:dialog_system}).

\subsection{Hash-Atlas Network} \label{subsec:hash-atlas}

In this section, we introduce a simple approach that directly maps images from various views of a scene onto 2D atlases, which relocates the 3D scene editing process to be executed in the 2D space. 
Similar techniques are initially employed for mapping video frames to atlases~\cite{kasten2021layered-atlas1,geyer2023tokenflow-atlas2,bar2022text2live-atlas3}, requiring continuous frames and smooth camera motion that is different with 3D scene data used in this paper. 
To achieve the desired editing capabilities outlined in this article, the atlases should meet the following criteria:
(1) Preventing excessive distortion and skew in the atlas for maintaining the visual model's comprehension. (2) The foreground and background atlases should be approximately aligned for accurate editing.
(3) A faster and more precise mapping is required to facilitate efficient editing.

\begin{figure*}[t]
    \centering
    \includegraphics[width=1.\textwidth]{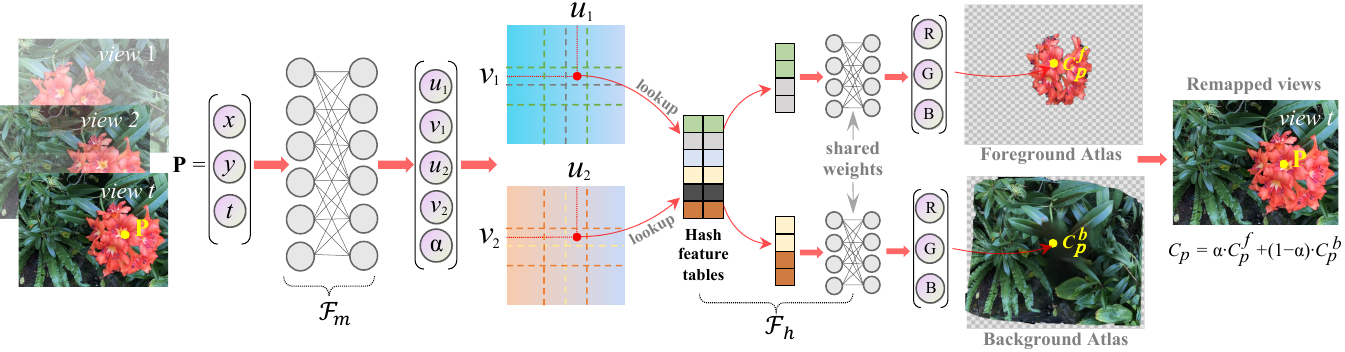}
    \caption{\textbf{Illustration of Hash-Atlas}. Given the views of a 3D scene, $F_m$ first maps each pixel coordinate from the views to two UV spaces and predicts the transparency of the foreground atlas. Subsequently, $F_h$ predicts the RGB values at each coordinate in the UV space, thereby obtaining the foreground and background atlases. When the atlases are edited, they can be mapped back to the original scene views.
    }
    \label{fig:hash_atlas}
\end{figure*}

\noindent{\textbf{Hash-Atlas Formulation.}}
To meet the aforementioned conditions, we devise a network based on a hash structure~\cite{muller2022instant}, as illustrated in Fig.~\ref{fig:hash_atlas}.
Assuming there are $T$ views in the scene, the point $P=[x, y, t]$ at the $t$-th view is mapped to two different UV coordinates using the function $F_m$:
\begin{equation}
    [u_1, v_1, u_2, v_2, \alpha] = F_m(x, y, t),
\label{eq:p2uv}
\end{equation}
where $(u_1, v_1)$ and $(u_2, v_2)$ represent the coordinates in the two UV spaces. The parameter $\alpha$, ranging between 0 and 1, signifies the weight of the pixel value in the foreground atlas.
Then $F_h$ is used to predict the RGB values corresponding to the foreground and background atlases in the UV coordinate:
\begin{equation}
    C_p^f = F_h(u_1, v_1); \quad C_p^b = F_h(u_2, v_2),
\label{eq:uv2atlas}
\end{equation}
where $F_h$ incorporates a hash structure~\cite{muller2022instant} to capture texture details in images and enable faster model training and inference. 
Furthermore, to share the weights of $F_h$ for two different UV coordinates, $(u_1, v_1)$ is arranged to the $[0, 0.5]$ interval, and $(u_2, v_2)$ to the $[0.5, 1]$ interval.

After obtaining the pixel values $C_p^f$ and $C_p^b$ in atlases, the original pixel value in the scene's view for point $P$ can be reconstructed as follows:
\begin{equation}
    C_p = \alpha \cdot C_p^f + (1-\alpha) \cdot C_p^b.
\label{eq:rec_p}
\end{equation}
When atlases are edited, each view of the 3D scene with editing effects can be restored by Eq.~\ref{eq:rec_p} without retraining the Hash-Atlas network again. 

\vspace{1mm}
\noindent{\textbf{Training and Loss Terms.}}
To ensure that the obtained atlases appear more natural, avoiding excessive tilting and distortion of objects, during the early stages of model training, we exclusively employ the $P=[x,y,0]$ from the $0$-th view. 
Then, the pre-training positional loss is defined as follows:
\begin{equation}
    \mathcal{L}_{pos} = ||x-u_1|| + ||x-u_2|| + ||y-v_1|| + ||y-v_2||.
\end{equation}
This loss encourages minimal changes in the position of the scene from the $0$-th view after the coordinate mapping.

Moreover, The pre-training for \(\alpha\) involves initially determining what the foreground of the scene is by a VQA model~\cite{li2022blip} and its corresponding mask by a  
segmentation model~\cite{kirillov2023sam, liu2023grounding-dino}. Assuming the foreground mask for point \(P\) is \(m_p\), the loss for pre-training \(\alpha\) is defined as follows:
\begin{equation}
    \mathcal{L}_{\alpha} = \text{CE}(\alpha_p, m_p) + ||(1-\alpha_p) C_{p}^{f}||,
\end{equation}
where \(\text{CE}\) represents the cross-entropy loss. The second term encourages the \(\alpha\) and the foreground atlas to be sparse, which facilitates a distinct separation between the contents of the foreground and background atlases.

Upon completing pre-training, the entire model can be trained by supervising the reconstruction view from the atlases.
However, we observe that such training leads to noticeable region omissions in the background atlas, which negatively impacts subsequent editing tasks.
To address this issue, we introduce the inpainting loss. Initially, we utilize the ProPainter model~\cite{zhou2023propainter} to perform preliminary inpainting on the masked background, generating a new set of inpainted views. Assuming that the point \(P\) in the original view corresponds to the \(\tilde{P}\) in the inpainted view, the reconstruction loss can be expressed as follows:
\begin{equation}
    \mathcal{L}_{rec} = \mathcal{L}_{rec}^{ori} + \mathcal{L}_{rec}^{pro} = || C_p - GT(C_p) || + ||C_{\tilde{p}}^b - GT(C_{\tilde{p}}^b) ||,
\end{equation}
where \(GT(\cdot )\) denotes the ground truth obtained from the original or inpainted views of the scene. 
In addition, following~\cite{kasten2021layered-atlas1}, we incorporate rigid and flow constraints on the scene. 
The purpose of the $\mathcal{L}_{rigid}$ is to maintain the relative spatial location between different points without drastic changes. Simultaneously, the $\mathcal{L}_{flow}$ encourages mapping the corresponding points from different views to the same location on the atlas. Therefore, the total loss can be expressed as follows:
\begin{equation}
    \mathcal{L}_{total} = \mathcal{L}_{init} + \mathcal{L}_{rec}^{ori} + \mathcal{L}_{rec}^{pro} + \mathcal{L}_{rigid} + \mathcal{L}_{flow},
\end{equation}
where $\mathcal{L}_{init} = \mathcal{L}_{pos} + \mathcal{L}_{\alpha}
$ is only employed during the initial training phase.

\subsection{Editing in Atlas Space}\label{subsec:editing_atlas}
We observe that directly editing two atlases and then mapping them back to the scene views usually does not yield satisfactory editing results. This is mainly because a single atlas contains incomplete scene information, especially in sparse foreground atlas. This limitation prevents the editing model from acquiring complete scene semantics to always achieve reliable editing. Therefore, we design a merge-split strategy for editing atlases.
In this process, we leverage ChatGPT's parsing and a VQA model to identify the editing areas. If the areas involve foreground content, we overlay the foreground atlas on the background, serving it as the actual atlas for editing. Subsequently, we use the original foreground mask and the new object mask to separate the edited atlas. We use ``Executor'' to represent the actual editing process, as shown in Fig.~\ref{fig:overview_pipe}.

\subsection{Dialog System} \label{subsec:dialog_system}
\noindent{\textbf{Sensitivity to Scene Names.}}
As a language model, ChatGPT cannot directly access information outside of text.
However, given the multitude of files involved in the editing process, it is impractical to input all of them as text into ChatGPT. 
Consequently, we represent the involved files with a single string formatted as `xxx.scn'. This string is unique and meaningless to prevent ChatGPT from fabricating scene names.
Although this scene name is not a genuinely readable file, further processing by the Frontend and Backend allows \modelname{} to handle real files effectively. The Frontend organizes the edited results and ChatGPT's outputs into user replies, while the Backend distributes the real scene files involved in the editing process and manages the new scene's names and files.

\vspace{1mm}
\noindent{\textbf{Reasoning for User Queries.}}
When presented with user input, ChatGPT simulates a thought process: ``Do I need to use a visual tool?'' → ``Which tools do I need?'' → ``What should be the specific input for the tools?''. Therefore, it is essential to pre-inject information about each visual expert into ChatGPT to complete this reasoning process. Similar to \cite{wu2023visual-gpt,yang2023mm-react}, we annotate each visual tool with four categories: the tool's name, under what circumstances it should be used, the required parameters, and specific input examples.

\begin{table}[t]
    \centering
    \caption{\textbf{Quantitative comparisons with LNA~\cite{kasten2021layered-atlas1} for atlas}. We report four metrics that are averaged over various scenes from different datasets, achieving a 1.1$\sim$3.2dB improvement in PSNR, a 14.2$\sim$18.6x acceleration in training, and a 7.6$\sim$9.0x increase in FPS. Thus, our method provides higher-quality atlases for the subsequent editing process and ensures faster mapping from atlases to scene views.
    }
     \resizebox{0.95\textwidth}{!}{
    \begin{tabular}{c|cc|cc|cc|cc}
    \toprule
          & \multicolumn{2}{c|}{\textbf{LLFF}} & \multicolumn{2}{c|}{\textbf{TanksAndTemples}} & \multicolumn{2}{c|}{\textbf{IBRNet-collect}} & \multicolumn{2}{c}{\textbf{CE3D-collect}} \\
\cmidrule{2-9}          & LNA   & Hash-Atlas & LNA   & Hash-Atlas & LNA   & Hash-Atlas & LNA   & Hash-Atlas \\
    \midrule
    PSNR$\uparrow$ & 23.04  & \textbf{24.66} & 19.24 & \textbf{20.4} & 25.56 & \textbf{28.53} & 23.96 & \textbf{27.24} \\
    SSIM$\uparrow$ & 0.67  & \textbf{0.79 } & 0.57  & \textbf{0.61} & 0.77  & \textbf{0.87} & 0.63  & \textbf{0.84} \\
    LPIPS$\downarrow $ & 0.40  & \textbf{0.17 } & 0.59  & \textbf{0.46} & 0.31  & \textbf{0.11} & 0.53  & \textbf{0.12} \\
    Train Time (h)$\downarrow $ & 10.75  & \textbf{0.60 } & 14.69 & \textbf{1.03} & 9.75  & \textbf{0.56} & 14.02 & \textbf{0.75} \\
    Inference FPS$\uparrow $ & 0.66  & \textbf{5.93} & 0.25  & \textbf{2.11} & 0.79  & \textbf{6.13} & 0.67  & \textbf{5.57} \\
    \bottomrule
    \end{tabular}%
}
    \label{tab:compare_atlas}
    \vspace{-0.0cm}
\end{table}

\section{Experiments and Analysis} \label{sec:exp}
\begin{figure}[t]
    \centering
    \includegraphics[width=0.9\textwidth]{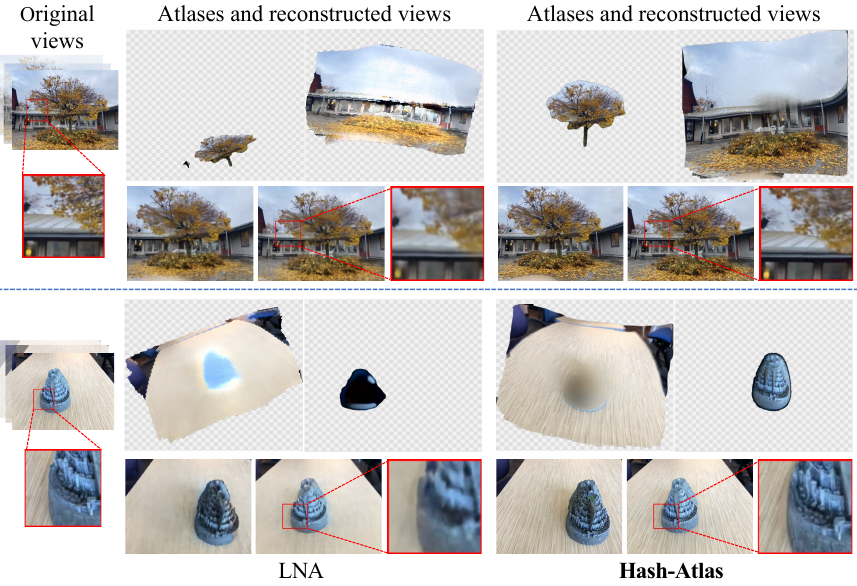}
    \caption{\textbf{Qualitative comparisons with LNA~\cite{kasten2021layered-atlas1} for atlas}. The atlases obtained from the Hash-Atlas preserve a greater amount of scene details.
    Moreover, our method ensures that the relative position of the objects within the foreground and background atlases remains largely invariant and guarantees that the scene objects in the atlases undergo minimal distortion and deformation.
    This provision of more precise scene target locations and more commonsensical visual information to subsequent editing models enables the attainment of better editing outcomes.
    }
    \label{fig:compare-atlas}
\end{figure}

We first quantitatively (Table~\ref{tab:compare_atlas}) and qualitatively (Fig.~\ref{fig:compare-atlas}) show the high-quality performance of our Hash-Atlas network.
Then, we display the powerful multi-round dialogue and diverse editing abilities of the proposed \modelname{} (Fig.~\ref{fig:shocking} and~\ref{fig:chat-1}), as well as comparisons with other scene editing methods (Figs.~\ref{fig:compare_editing_text_multi} and~\ref{fig:compare-multi-round}).
We finally provide extensive ablation studies to validate our design choices (Fig.~\ref{fig:ablation} and Table~\ref{tab:ablation}) and limitations (Fig.~\ref{fig:limitations}) of our method. 
We encourage the reader to view supplementary materials and video demos for more experimental results.

\subsection{Implementation Details}
The datasets used in this study include LLFF~\cite{mildenhall2019llff}, NeRF-Art~\cite{wang2022nerfart}, IN2N-collect~\cite{haque2023instruct}, IBRNet-collect~\cite{wang2021ibrnet}, TanksAndTemple~\cite{knapitsch2017tanks}.
We additionally collect dozens of outdoor forward-facing datasets, denoted as \modelname{}-collect, to supplement scene types (See supplementary materials for more details).
We implement the Hash-Atlas network using PyTorch, with the optimizer as AdamW, and the learning rates of $F_m$ and $F_h$ are 1e-3 and 1e-2, respectively. 
We utilize the ``gpt-3.5-turbo'' API to access ChatGPT. 
The 3D models available for invocation are TensoRF~\cite{chen2022tensorf} and Gaussian-Splatting~\cite{kerbl20233dgaussian}. 
The 2D visual experts including VQA~\cite{li2022blip},
Image Generation~\cite{rombach2022high},
Image Caption~\cite{li2022blip},
Super-Resolution~\cite{rombach2022high}
Text-driven Stylize~\cite{brooks2022instructpix2pix},
Image-driven Stylize~\cite{li2023blip2},
Segmentation~\cite{kirillov2023sam}.
In addition, we employ the models that transfer the image to Line~\cite{gu2022-img2line}, Edge~\cite{xu2017img2edge}, Hed~\cite{xie2015-img2hed-img2sketch}, Depth~\cite{ranftl2021-img2depth-normal-2}, Normal Map~\cite{ranftl2021-img2depth-normal-2}, Sketch~\cite{xie2015-img2hed-img2sketch}, Pose\cite{cao2017img2pose}, and also the reverse conversion using ControlNet~\cite{zhang2023controlnet}.
All editing experiments are conducted on two A800 80GB GPUs.

\subsection{Atlas Reconstruction Results}
A high-quality atlas representation is fundamental to effective scene editing. Given the scarcity of current methods for mapping 3D scene views onto atlases, we benchmark our proposed approach against LNA~\cite{kasten2021layered-atlas1}, a technique for constructing atlases in video, to assess the quality and adaptability of the generated atlases for 3D scene editing.
The results, as presented in Table~\ref{tab:compare_atlas} and Fig.~\ref{fig:compare-atlas}, reveal that the proposed Hash-Atlas significantly outperforms LNA and exhibits superior adaptability to the subsequent 3D scene editing operations.

\begin{figure*}[htbp]
    \centering
\includegraphics[width=1.\textwidth]{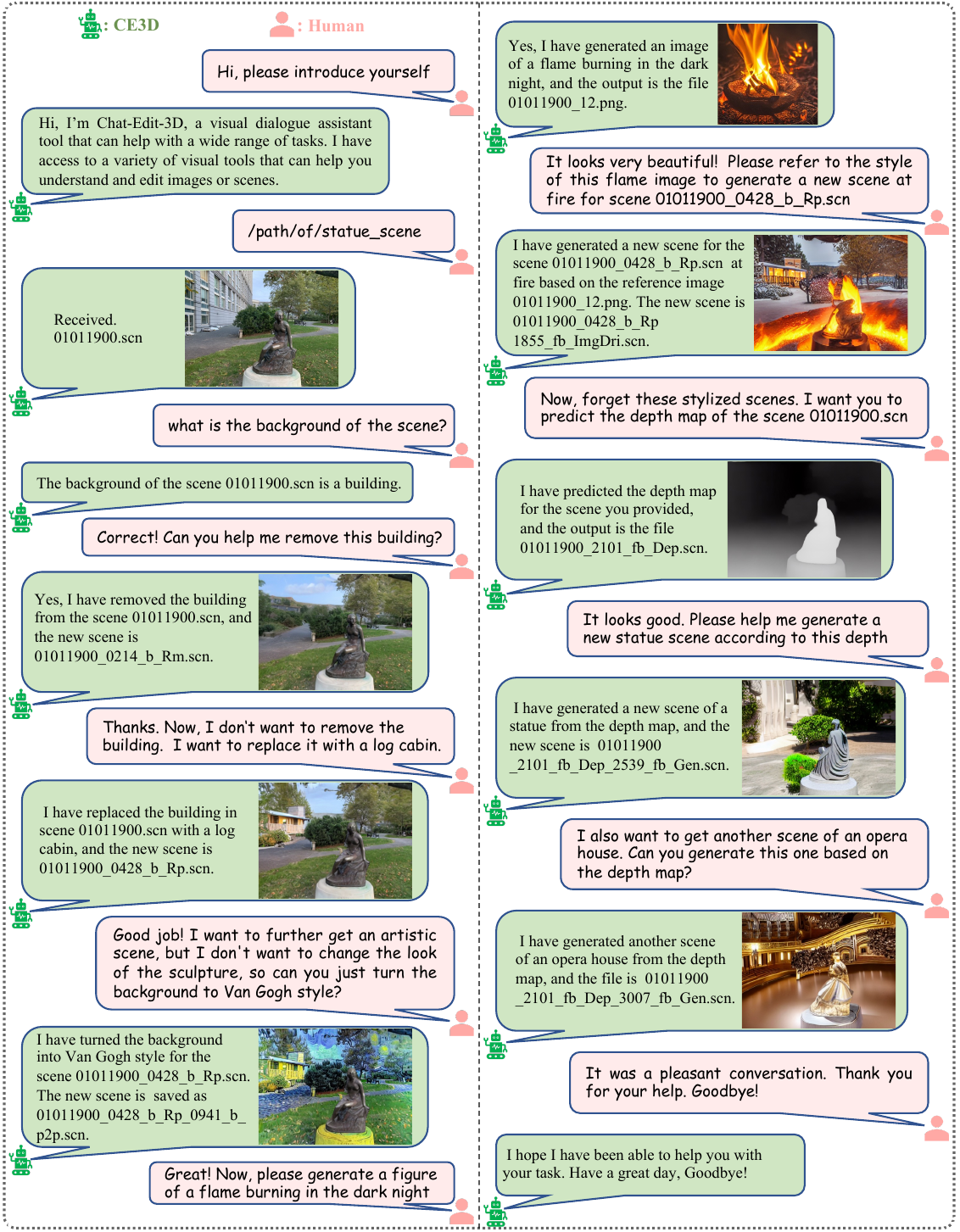}
    \caption{\textbf{Multi-round dialogue editing results}. \modelname{} performs editing of 3D scenes on the 2D atlases, thus facilitating seamless compatibility with various existing 2D models and boasting multiple editing capabilities. Moreover, by utilizing LLMs for the management of various visual models and parsing user text inputs, \modelname{} enables users to engage in more flexible text inputs and easily achieve multi-round dialogue.
    }
    \label{fig:chat-1}
\end{figure*}

\begin{figure}
    \centering
    \subfloat[\textbf{Comparisons for various editing abilities}. State-of-the-art methods show limited editing performance. In contrast, the proposed \modelname{} offers facile compatibility with various 2D visual models, thereby enabling the integration of a diverse array of editing functionalities.]{
        \includegraphics[width=0.95\textwidth]{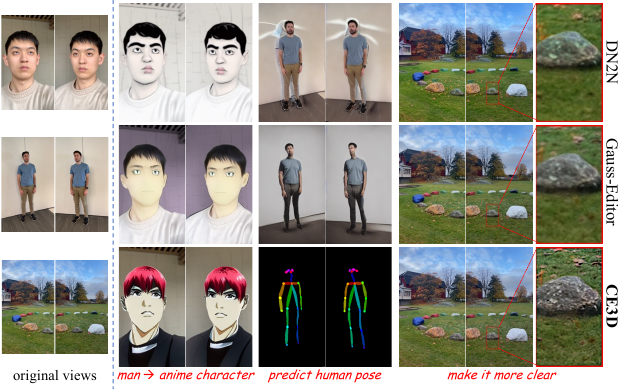}
        \label{fig:compare_editing}
    }
    \vspace{0.3cm}
    \hfill
    \subfloat[\textbf{Comparisons for text-query diversity.} We show the qualitative results for the same editing type with three different text queries. Although all three editing instances involve edge detection, IN2N produces notably varied results when the accompanying text varies. While \modelname{} possesses a robust language comprehension capability that mitigates instability in the editing outcomes resulting from textual discrepancies.
    ]{
        \includegraphics[width=0.95\textwidth]{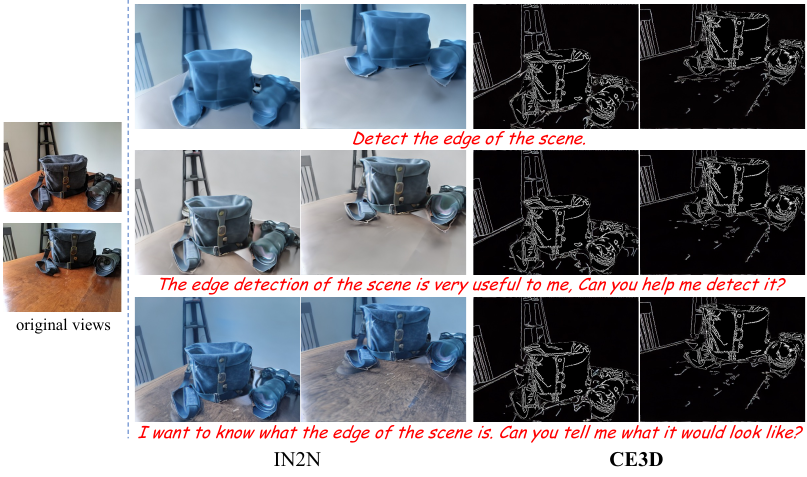}
        \label{fig:compare_text_multi}
    }
    \caption{Comparison with other editing methods for editing abilities and text-query diversity.}
    \label{fig:compare_editing_text_multi}
\end{figure}

\begin{figure}[t]
  \centering
  \includegraphics[width=0.9\textwidth]{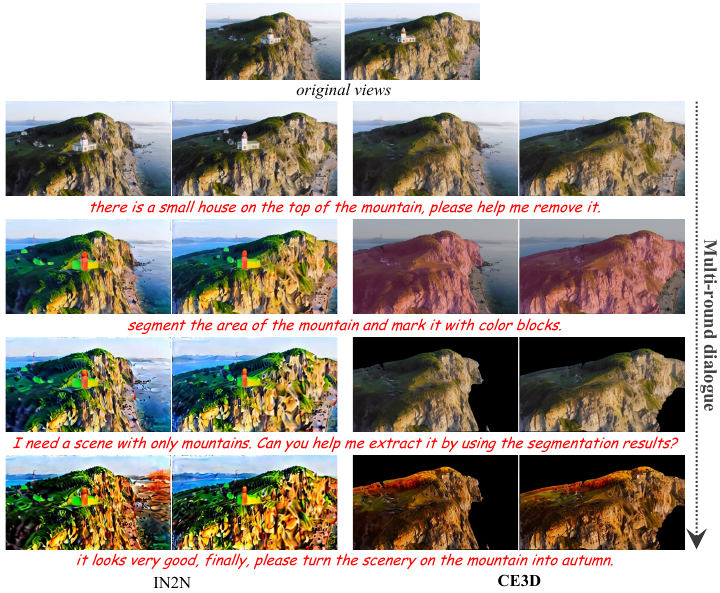}
  \caption{\textbf{Comparisons with IN2N for multi-round dialogue}. IN2N lacks sufficient editing capabilities and falls short in text parsing, making it challenging to handle complex multi-round dialogue tasks. By introducing a decoupled novel editing framework, our method effectively addresses these challenges and outperforms IN2N significantly.}
  \vspace{-2pt}
  \label{fig:compare-multi-round}
\end{figure}

\subsection{3D Scenes Editing Results}
\noindent{\textbf{Editing Cases of Multiple Rounds Dialogue.}}
In Fig.~\ref{fig:chat-1}, we present a 12-round dialogue example where users provide free-form text queries, and \modelname{} responds with user text, images, or edited results. 
These dialogues involve various editing types, such as object removal or replacement, text-driven or image-driven style transfer, depth map prediction, and scene regeneration based on text and depth map conditions. Furthermore, it can also accomplish tasks like VQA related to a scene and basic text-based dialogues. Due to space constraints, this example only partially demonstrates \modelname{}'s capabilities. We present a broader array of editing outcomes in the supplementary materials.

\vspace{1mm}
\noindent{\textbf{Comparisons.}}
Fig.~\ref{fig:compare_editing} presents a comparison of editing results between \modelname{} and baseline methods, which demonstrate the superior and more comprehensive scene editing capabilities of \modelname{}. In Fig.~\ref{fig:compare_text_multi}, we showcase \modelname{}'s advantages in handling diverse text queries and its robustness.
We further compare the editing results of 4-round dialogue with IN2N in Fig~\ref{fig:compare-multi-round}, which attests to \modelname{}'s stable engagement potential in interactive dialogue-based editing processes.

\begin{table}[htb]
    \centering
    \caption{\textbf{Quantitative ablation studies of \modelname{}}. Here ``-'' indicates that the model variant fails to provide edited results. The CLIP calculates the similarity between edited results and text. CDS means CLIP Directional Score~\cite{haque2023instruct}, which measures how much the change in text captions agrees with the change in images.}
    \setlength\tabcolsep{3pt}
    \resizebox{1.\textwidth}{!}{
    
    \begin{tabular}{c|ccccc|ccccc}
    \toprule
          & \multicolumn{5}{c|}{\textbf{LLFF}}    & \multicolumn{5}{c}{\textbf{CE3D-collect}} \\
\cmidrule{2-11}          & w/o $\mathcal{L}_{init}$ & w/o $\mathcal{L}_{rec}^{pro}$ & w/o SSN & w/o Excutor & \textbf{Full} & w/o $\mathcal{L}_{init}$ & w/o $\mathcal{L}_{rec}^{pro}$ & w/o SSN & w/o Excutor & \textbf{Full} \\
    \midrule
    CLIP$\uparrow$ & 0.242 & 0.267 & $-$   & 0.190 & \textbf{0.304} & 0.286 & 0.263 & $-$   & 0.249 & \textbf{0.352} \\
    CDS$\uparrow$ & 0.163 & 0.179 & $-$   & 0.154 & \textbf{0.192} & 0.214 & 0.197 & $-$   & 0.201 & \textbf{0.248} \\
    \bottomrule
    \end{tabular}%

    }
    \label{tab:ablation}
     \vspace{10pt}
\end{table}

\begin{figure}[!htb]
    \centering
    \subfloat[\textbf{Loss ablation on the Hash-Atlas network.} The text query is ``remove sofa''.]{
        \includegraphics[width=0.95\textwidth]{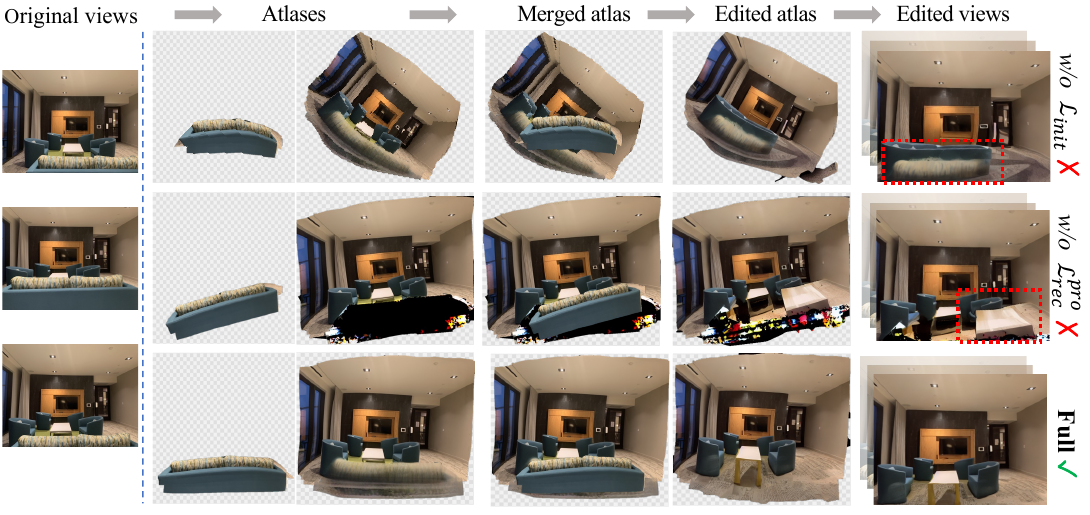}
        \label{fig:ablation_hash}
    }
    \vspace{0.5cm}
    \hfill
    \subfloat[\textbf{Designed component ablation on the editing strategy and dialogue system.}
    ]{
        \includegraphics[width=0.95\textwidth]{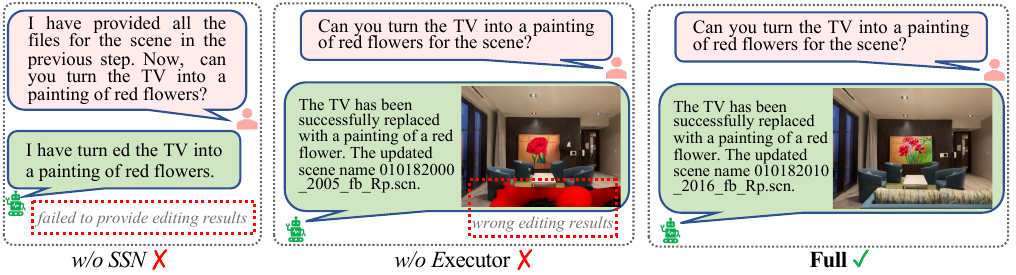}
        \label{fig:ablation_llm}
    }
    \caption{\textbf{Qualitative ablation studies of \modelname{}}. Red box marks failed results.}
    \label{fig:ablation}
    \vspace{2pt}
\end{figure}

\subsection{Ablation Studies} \label{subsec:ablation}
In Fig.~\ref{fig:ablation}, we illustrate the significance of each design element incorporated in our method. 
The absence of loss $\mathcal{L}_{init}$ and $\mathcal{L}_{rec}^{pro}$ leads to decreased quality and rationality in the edited results. 
The SSN (Sensitivity to Scene Names) undertakes the input and output management of scene names and related files. Removing it and directly using the original file paths as inputs would cause \modelname{} to fail to complete edits correctly and provide users with reasonable responses.
The Executor highlights the importance of managing the merging and separation between foreground and background atlases when editing the atlases. Editing the two atlases independently without the use of the Executor may lead to erroneous editing outcomes due to the lack of complete scene information.
We also conduct editing on scenes from  LLFF and CE3D-collect datasets, and quantitatively evaluated the editing results, as shown in Table~\ref{tab:ablation}, which further demonstrate the importance of each designed component in our method for achieving high-quality and high-reliability dialogue-based editing.
\begin{figure}[t]
    \centering
\includegraphics[width=1.\textwidth]{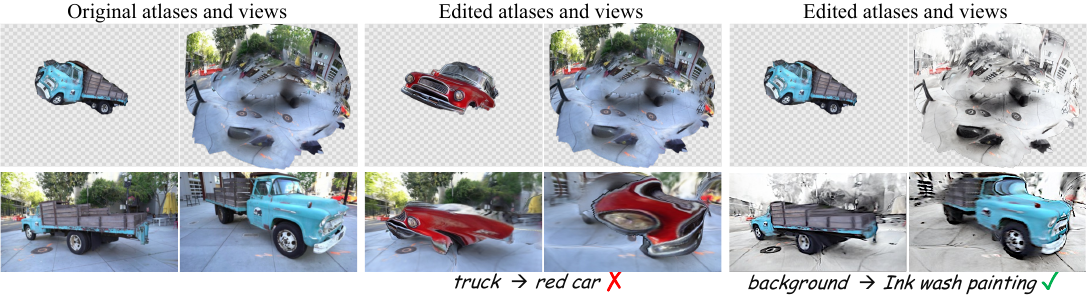}
    \caption{\textbf{Limitations}. The distorted atlases in 360-degree scenes may not integrate well with 2D visual models, potentially leading to some instances of failed editing.
    }
    \label{fig:limitations}
    \vspace{2pt}
\end{figure}

\subsection{Discussions and Limitations}
While \modelname{} demonstrates the potential for conversational editing in 3D scenes, it still has some limitations. These include a reliance on the correctness of LLMs for text query parsing and the need for manual tuning of prompts for various visual models. The effectiveness of scene editing is also dependent on the performance of the visual experts themselves. Our method supports limited geometry editing, such as object removal and replacement. However, editing location, scale, and orientation are not handled in this paper. Additionally, the atlas obtained on 360-degree scenes often exhibit severe distortion and the training data for existing 2D visual models lacks such atlas modality, potentially leading to unreasonable or unsatisfactory editing outcomes (Fig.~\ref{fig:limitations}).

\section{Conclusions} \label{sec:conclusion}
Our work addresses the limitations of previous efforts that coupled 3D scene representation with 2D image editing, as well as the challenges they face in enabling more intelligent interactive designs. We demonstrate that representing scenes as 2D atlases (mapping from 3D scene views to 2D plane images) conveniently facilitates compatibility between 3D scene editing and any 2D visual model. Through the automatic management of LLMs, we endow \modelname{} with rich editing effects and sustainable conversational editing capabilities. 
Although there remains considerable room for further advances in atlas optimization and editing in complex 360-degree scenes, this work makes significant contributions to interactive editing with real-world scenes, where scene editing is no longer confined to a limited set of visual models and can support a more flexible and unconstrained editing paradigm.

%
%
\bibliographystyle{splncs04}
\bibliography{mainbib}

\clearpage

\begin{center}
    \Large
    \vspace{50mm}
     \textcolor[rgb]{0.,0.18,0.85}{\textbf{Supplementary Materials}}
    \\[18pt]
    \normalsize 
\end{center}

\section{Additional Implementation Details}
\subsection{Training for Hash-Atlas Network}
In the Hash-Atlas network structure showcased in Fig.~4 in the main manuscript, we establish $F_m$ as an eight-layer MLP with a width of 256, along with ReLU activations. 
For $F_h$, we set the 
base resolution, number of levels, per-level scale, hash table size, and feature dimensions to 
16, 16, 1.5, $2^{15}$, and 2 respectively.  The MLP, followed by the hash tables, has two hidden layers with a width of 64, The activation function is ReLU for each hidden layer and tanh for the output layer of color.
The generated atlases' resolution stands at $1000\times 1000$ pixels.
We implement our method using the PyTorch framework and AdamW Optimizer. The initial learning rates for $F_m$ and $F_h$ are 1e-3 and 1e-2, respectively. We run 100,000 steps with a batch size of 10,000 pixel points.
The $\mathcal{L}_{pos}$ loss is utilized solely for the initial 1,000 iterations, after which the $\mathcal{L}_{\alpha}$ loss is trained only for the first 30,000 iterations. 

\subsection{Prompts for CE3D}
\noindent\textbf{Prompt for Sensitivity to Scene Names.} We present this prompt in Table~\ref{tab:prompt_ssn}. The scene name is associated with relevant scene files, encompassing various original view images of the scene, camera poses, foreground masks, atlases, and intermediate files produced during the editing process.
Upon perceiving the scene name, ChatGPT triggers the actual file orchestration executed by Frontend or Backend, as discussed in Sec 3.3 of our main text.

\begin{table}[!ht]
\centering
\caption{Example prompts for sensitivity to scene names.}
\vspace{-0.4cm}
\begin{mdframed}[backgroundcolor=white]
    \resizebox{0.95\textwidth}{!}{
    \begin{tabular}{p{0.8\linewidth}}
       \textit{As a language model, Chat-Edit-3D can not directly read scenes, but it has a list of tools to finish different visual tasks. Each scene will have a file name formed as ``xxx.scn'', and the file name has no meaning. When talking about scenes, Chat-Edit-3D is very strict with the file name and never fabricates nonexistent files. Chat-Edit-3D can invoke different tools to indirectly understand and edit scenes. Chat-Edit-3D is able to use tools in a sequence and is loyal to the tool observation outputs rather than faking the scene content and file name. It will remember to provide the file name from the last tool observation.}
\end{tabular}}
\end{mdframed}
\label{tab:prompt_ssn}
\end{table}
\begin{table}[!ht]
\centering
\caption{Example prompts for visual experts.}
\vspace{-0.5cm}
\begin{mdframed}[backgroundcolor=white]%
    \resizebox{0.95\textwidth}{!}{
    \begin{tabular}{p{0.8\linewidth}}
     \textbf{Tool Name}: \textit{Generate a New Scene Base on a Reference image and Text prompt.}\\
     \textbf{Description}: \textit{useful when you want to generate a new scene from both the user description and a reference image. Like: generate a new real scene of an object or something for scene X based on the reference image Y, or refer to image Y to generate a new scene at night for image X. The input to this tool should be a comma-separated string of three, representing the scene name, reference image path, and the user description.}
\end{tabular}}
\end{mdframed}
\label{tab:prompt_expert}
\end{table}

\noindent\textbf{Role Customization of Visual Experts.}
\modelname{} enables convenient integration of existing 2D and 3D visual experts. For LLMs to effectively leverage these experts, customization of their roles and specification of parameters are required. We provide an illustrative example of such customization, as depicted in Table~\ref{tab:prompt_expert}.
Here, certain visual experts necessitated the use of multiple visual models. For instance, replacing an object requires employing a segmentation model to remove the target object and using an inpainting model to fill it with a new object. In such cases, we also consider the replacement operation as a custom-designed visual expert.

\section{Additional Experiments}
\begin{figure}[t]
    \centering
    \includegraphics[width=\textwidth]{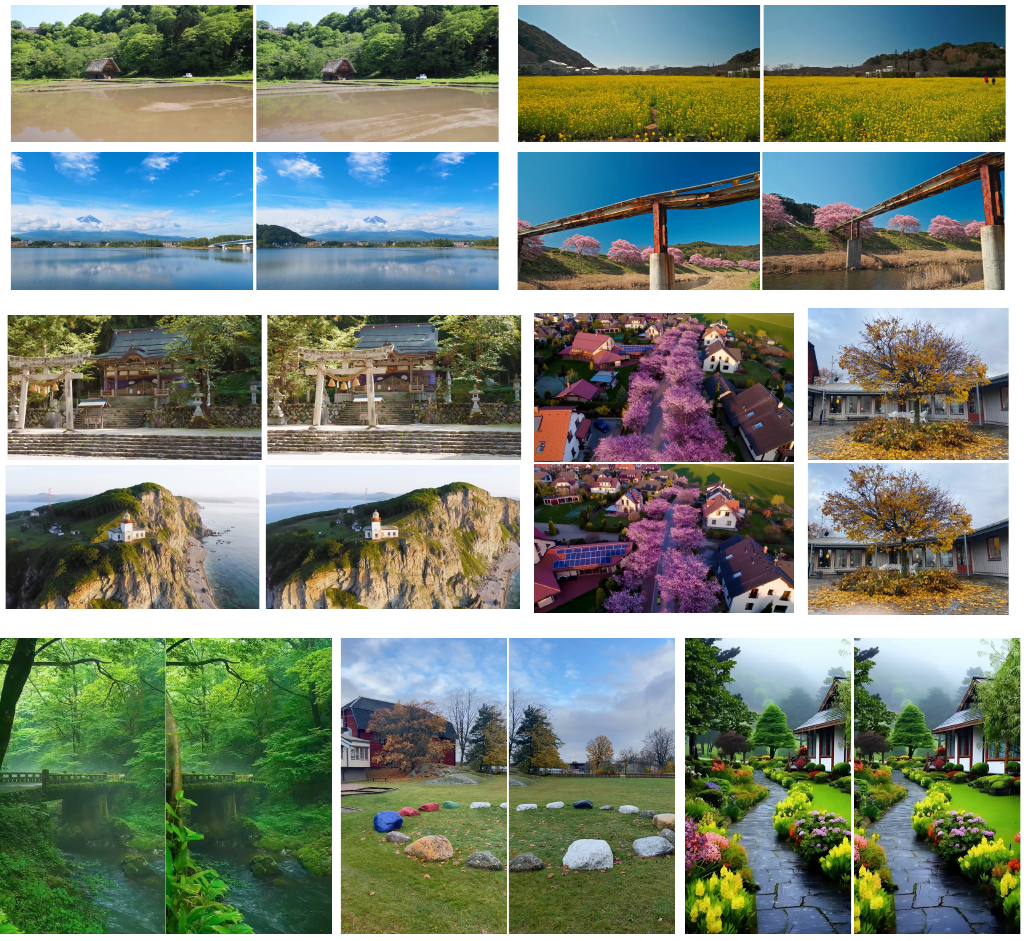}
    \caption{\textbf{Examples of the CE3D-Collect dataset.}
    }
    \label{fig:ce3d-dataset}
\end{figure}

\noindent\textbf{{CE3D-Collect Dataset.}}
Given that the current forward-facing scenarios predominantly involve indoor objects, we additionally collect forward-facing datasets encompassing outdoor expansive viewpoints, as illustrated in Fig.~\ref{fig:ce3d-dataset}. The resolution range of the data spans from 1K to 2K, and we employ COLMAP~\cite{schoenberger2016colmap1,schoenberger2016mvs-colmap2} to estimate the camera poses for each dataset.

\begin{figure}[t]
    \centering
    \includegraphics[width=\textwidth]{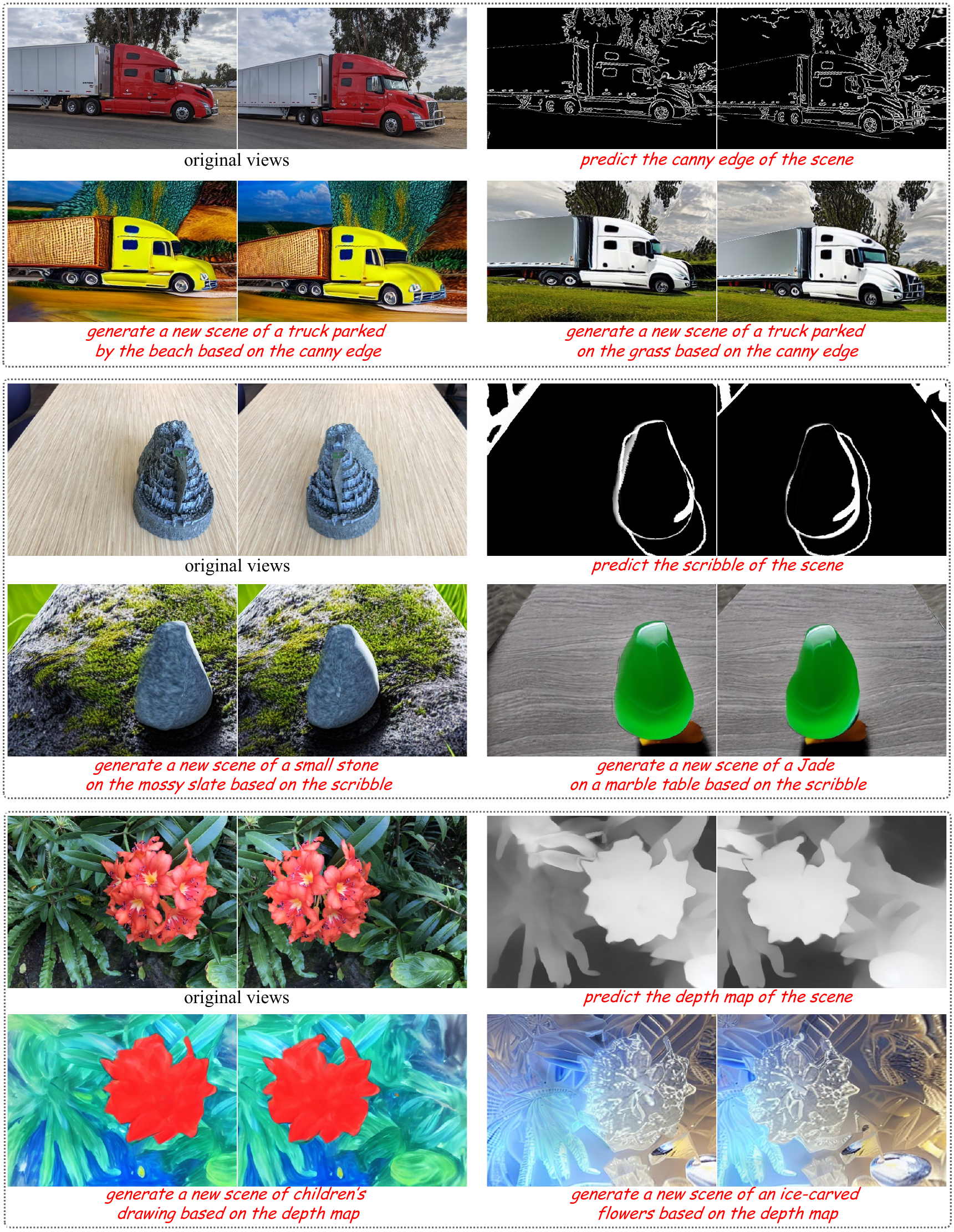}
    \caption{\textbf{Predict the canny edge, scribble and depth map of scenes, and generate new scenes based on these conditions and text prompts.}
    }
    \label{fig:edge-hed-scribble}
\end{figure}

\begin{figure}[t]
    \centering
    \includegraphics[width=\textwidth]{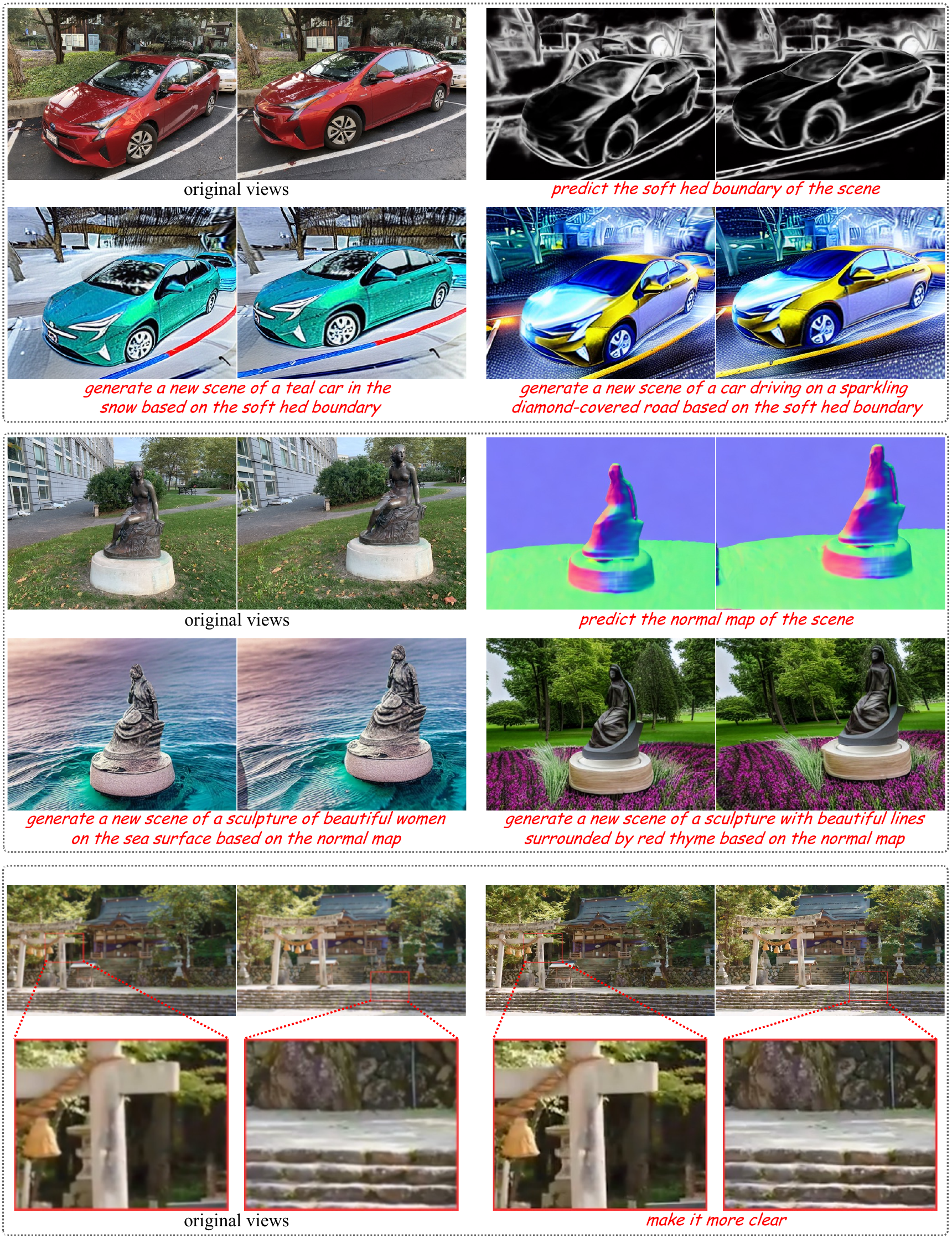}
    \caption{\textbf{Predict the soft hed boundary and normal map of scenes, and generate new scenes based on these conditions and text prompts. Also, perform super-resolution on a scene.}
    }
    \label{fig:normal-hed}
\end{figure}

\begin{figure}[t]
    \centering
    \includegraphics[width=\textwidth]{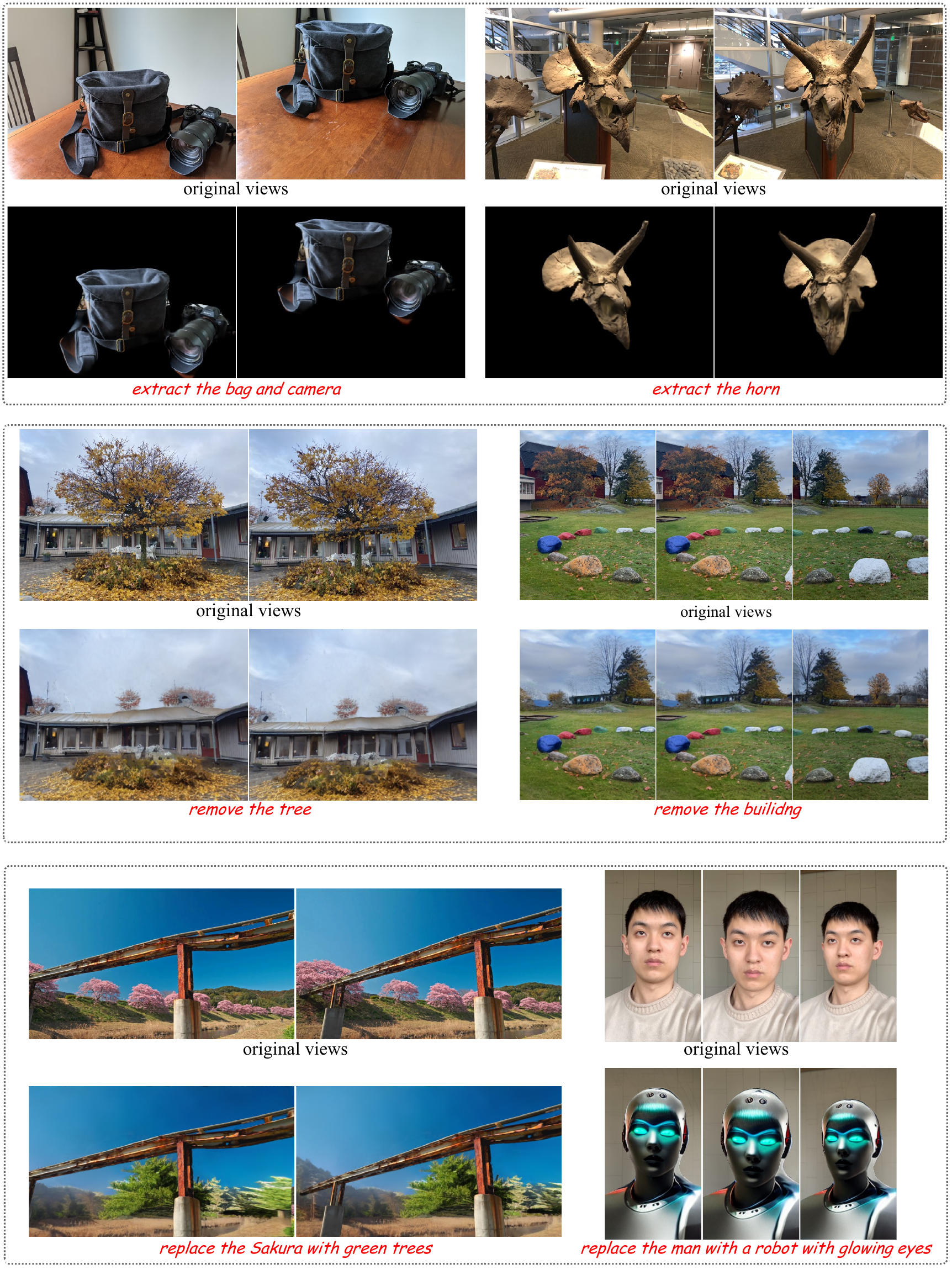}
    \caption{\textbf{Editing results of scene extraction, removal and replacement.}
    }
    \label{fig:extract-remove-replace}
\end{figure}

\begin{figure}[t]
    \centering
    \includegraphics[width=\textwidth]{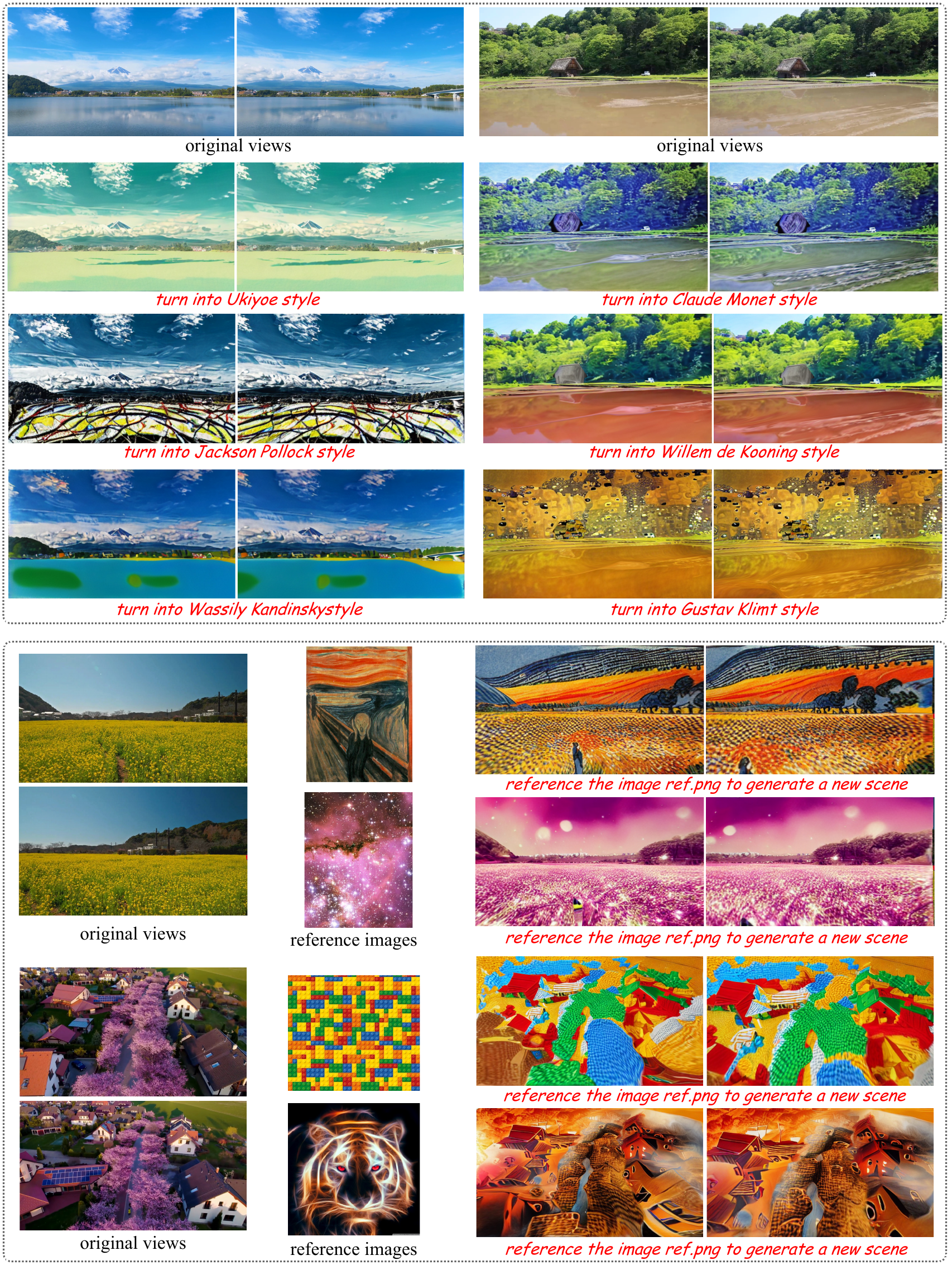}
    \caption{\textbf{Editing results of text-driven style transfer and image-driven style transfer.}
    }
    \label{fig:style}
\end{figure}

\noindent\textbf{More Editing Results.}
We present additional editing examples as shown in Figs~\ref{fig:edge-hed-scribble}, \ref{fig:normal-hed}, \ref{fig:extract-remove-replace} and \ref{fig:style}. CE3D transforms scene editing into operations in two-dimensional space, making it easier to integrate existing 2D editing models and achieve richer editing results.

\end{document}